\documentstyle[11pt]{article}

\title{Memoization in Constraint Logic Programming}
\author{Mark Johnson\thanks{
This paper was presented at the
First Workshop on Principles and Practice of
Constraint Programming, April 28--30 1993,
Newport, Rhode Island.
The research descrined in it was
initiated during a summer visit to the IMSV,
Universit\"at Stuttgart; which I would like to thank
for their support.
Thanks also to Pascal van Hentenryck and Fernando Pereira for
their important helpful suggestions.}\\
Brown University}
\date{}

\newtheorem{Theorem}{Theorem}

\newcommand{\clause}[2]{{#1 \leftarrow #2}}

\newcommand{\program}{\mbox{\sf program}}
\newcommand{\lookup}{\mbox{\sf table}}
\newcommand{\solution}{\mbox{\sf solution}}

\newcommand{\emptylist}{[]}
\newcommand{\nil}{\mbox{\sf nil}}

\newcommand{\uline}{{\tt \_}}

\newcommand{\eop}{ \rule{2mm}{2mm} }

\def\picture #1 by #2 (#3){
 \vbox to #2{
   \hrule width #1 height 0pt depth 0pt
   \vfill
   \includegraphics{#3}}}

\def\beginprolog{\par\vspace{-.5ex}\begingroup\leftskip=\leftmargini\sf
                 \setlength{\parindent}{0in}
                 \setlength{\parskip}{0in}
                 \def\par{\leavevmode\endgraf}  
                 \catcode`\^=12 
                 \catcode`\_=12 %
                 \catcode`\|=12 %
                 \obeylines%
                 \catcode`\ =\active%
                 \catcode`\%=12\catcode`\`=\active}
{\catcode`\ =\active\global\let =\enskip}
{\catcode`\`=\active\gdef`{\relax\lq}}

\def\endprolog{\endgroup\par\vspace{2.2ex}}

\begin{document}

\maketitle

\section{Introduction}
This paper shows how to apply memoization (caching of subgoals
and associated answer substitutions) in a constraint
logic programming setting.
The research is
is motivated by the desire to apply
constraint logic programming (CLP)
to problems in natural language processing.

In general,
logic programming provides an excellent theoretical
framework for computational linguistics~\cite{ps:pnlp}.
CLP extends ``standard'' logic programming by allowing
program clauses to include constraints from a
specialized contraint language.
For example, the CLP framework allows
the feature-structure constraints that have proven
useful in computational linguistics~\cite{s:intro}
to be incorporated into logic programming in a natural
way~\cite{cw:clogic,hs:drcl,gs:fcl}.

Because modern linguistic theories describe natural
language syntax as a system of
interacting ``modules'' which jointly
determine the linguistic structures
associated with an utterance~\cite{c:kol},
a grammar can be regarded as a conjunction
of constraints whose solutions are exactly the
well-formed or grammatical
analyses.
Parsers for such grammars typically coroutine between
a tree-building component that generates nodes of the parse
tree and the well-formedness constraints imposed by the linguistic modules
on these
tree structures~\cite{gkpc:prolog,mj:ukl,mj:pad}.
Both philosophically and practically, this fits in well
with the CLP approach.

But the standard CLP framework inherits some of the
weaknesses of the SLD resolution procedure that it is
based on.
When used with the standard
formalization of a context-free grammar
the SLD resolution procedure behaves as
a recursive descent parser.
With left-recursive grammars
such parsers typically fail to terminate
because a goal corresponding to a prediction
of a left-recursive category can reduce to an
identical subgoal (up to renaming), producing an
``infinite loop''.
Standard techniques for
left-recursion elimination~\cite{mthmy:bup,ps:pnlp}
in context-free grammars
are not always directly applicable to grammars formulated
as the conjunction of several constraints~\cite{mj:lcpt}.

With memoization, or the caching of intermediate goals (and their
corresponding answer substitutions),
a goal is solved only once and its solutions are
cached; the solutions to identical goals are obtained
from this cache.
Left recursion need not lead to non-termination because
identical subgoals are not evaluated, and the
infinite loop is avoided.
Further, memoization can sometimes provide the advantages
of dynamic programming approaches to parsing:
the Earley deduction proof procedure (a memoized version
of SLD resolution) simulates an Earley parse~\cite{e:ep}
when used with the standard formalization
of a context-free grammar~\cite{pw:ed}.

Thus constraint logic programming and memoization are two
recent developments in logic
programming that are important for natural language processing.
But it is not obvious how, or even if, the two can
be combined in a single proof procedure.
For example, both Earley Deduction~\cite{pw:ed} and
OLDT resolution~\cite{ts:oldt,wa:mem}
resolve literals in a strict left-to-right order,
so they are not capable of rudimentary constraint
satisfaction techniques such as goal delaying.
The strict left-to-right order restriction is
relaxed but not removed in~\cite{v:1,v:2},
where literals can be resolved in any local order.
This paper describes soundness and completeness
proofs for a proof procedure that extends
these methods to allow for goal delaying.
In fact, the lemma table proof procedure generalizes naturally to
constraint logic programming over arbitrary domains,
as described below.

The lemma table proof procedure generalizes Earley Deduction
and OLDT resolution in three ways.
\begin{itemize}
\item Goals can be resolved in any order (including non-local orders),
      rather than a fixed left-to-right or a local order.
\item The goals entered into the table consist of non-empty sets of literals
      rather than just single literals.
      These sets can be viewed as
      a single program literal and zero or more constraints that are
      being passed down into the subsidary proof.
\item The solutions recorded in the lemma table may contain
      unresolved goals.  These unresolved goals can be thought
      of as constraints that are being passed out of the subsidary
      proof.
\end{itemize}

\section{A linguistic example}
Consider the grammar fragment in Figure~\ref{fig:grammar}
(cf.\ also~\cite[pages 142--177]{gkpc:prolog}).
The {\sf parse} relation holds between a string and
a tree if the yield of the tree is the string to be
parsed and tree satisfies a well-formedness condition.
In this example, the well-formedness condition is
that the tree is generated by a simple context-free
grammar, but in more realistic fragments the
constraints are considerably more complicated.

\begin{figure}
{\sf
\begin{tabular}{lll}
\hspace{0.15in} & \multicolumn{2}{l}{parse(String, Tree) :- wf(Tree, s),
y(Tree, String, []). } \\
 &  & \\
 & \multicolumn{2}{l}{y(\uline-Word, [Word{\tt|}Words], Words). } \\
 & \multicolumn{2}{l}{y(\uline/[Tree1], Words0, Words) :- y(Tree1, Words0,
Words). } \\
 & \multicolumn{2}{l}{y(\uline/[Tree1,Tree2], Words0, Words) :- } \\
 & \multicolumn{2}{l}{ \hspace{0.25in} y(Tree1, Words0, Words1), y(Tree2,
Words1, Words). } \\
 &  & \\
 & wf(np-kim, np). &\mbox{\rm\footnotesize \% NP $ \rightarrow$ \em Kim} \\
 & wf(n-friend, n). &\mbox{\rm\footnotesize \% N $ \rightarrow$ \em friend} \\
 & wf(v-walks, v). &\mbox{\rm\footnotesize \% V $ \rightarrow$ \em walks} \\
 & wf(s/[Tree1, Tree2], s) :- wf(Tree1, np), wf(Tree2, vp).
&\mbox{\rm\footnotesize \% S $ \rightarrow $ NP VP} \\
 & wf(np/[Tree1, Tree2], np) :- wf(Tree1, np), wf(Tree2, n).
&\mbox{\rm\footnotesize \% NP $ \rightarrow $ NP N} \\
 & wf(vp/[Tree1], vp) :- wf(Tree1, v). &\mbox{\rm\footnotesize \% VP $
\rightarrow $ V} \\
\end{tabular} }
\caption{A grammar fragment} \label{fig:grammar}
\end{figure}

Trees are represented by terms.  A tree consisting
of a single pre-terminal node labelled $C$ whose
single child is the word $W$ is represented by the
term $C${\sf -}$W$.  A tree consisting of a root
node labelled $C$ dominating the sequence of
trees $T_1 \ldots T_n$ is represented by the term
$C${\sf /[}$T_1, \ldots, T_n${\sf ]}.

The predicate {\sf wf(}{\em Tree, Cat}{\sf)}
holds if {\em Tree} represents a well-formed parse tree
with a root node labelled {\em Cat} for the
context free grammar shown in the comments.
The predicate {\sf y}({\em Tree,  S0, S}) holds
if {\em S0--S} is a ``difference list'' representing
the yield of {\em Tree}; it
collects the terminal items in the familiar tree-walking fashion.
{}From this program, the following instances of {\sf parse}
can be deduced
(these are meant to approximate
possessive constructions like {\em Kim's friend's friend walks}).
\beginprolog \sf
parse([kim,walks], s/[np-kim,vp/[v-walks]]).
parse([kim,friend,walks], s/[np/[np-kim,n-friend],vp/[v-walks]]).
parse([kim,friend,friend,walks],
s/[np/[np/[np-kim,n-friend],n-friend],vp/[v-walks]]).
$\ldots$
\endprolog

The parsing problem is encoded as a goal as follows.
The goal consists of an atom with the predicate {\sf parse}
whose first argument instantiated to the string to be parsed
and whose second argument is uninstantiated.
The answer substitution
binds the second argument to the parse tree.
For example, an answer substitution for
the goal {\sf parse([Kim,friend,walks], Tree)}
will have the parse tree for the string {\em Kim friend walks}
as the binding for {\sf Tree}.

Now consider the problem of parsing using the program shown
in Figure~\ref{fig:grammar}.
Even with the variable {\sf String} instantiated, both of the
subgoals of {\em parse}
taken independently have an infinite number of
subsumption-incomparable answer substitutions.
Informally,
this is because there are an infinite number of trees
generated by the context-free grammar, and there are
an infinite number of trees that have any given non-empty string
as their yield. Because the standard memoization techniques
mentioned above all compute all of the answer
substitutions to every subgoal independently, they never
terminate on such a program.

However, the set of answer substitions that
satisfy both constraints is finite, because
the number of parse trees with the same yield with respect to this
grammar is finite.
The standard approach to parsing with such grammars
takes advantage of this by using a selection
rule that ``coroutines'' the goals {\sf wf} and
{\sf y}, delaying all {\sf wf} goals until the first
argument is instantiated to a non-variable.
In such a system, the {\sf wf} goals function as constraints that
filter the trees generated by the goals {\sf y}.

In this example, however, there is a second, related, problem.
Coroutining is not sufficient to yield
a finite SLD tree,
even though the number of refutations is finite.
Informally, this is because the grammar in Figure~\ref{fig:grammar}
is left recursive, and the search space for a recursive
descent parser (which an SLD refutation mimics with such a program)
is infinite.

Figure~\ref{fig:sld} shows part of an infinite SLD derivation
from the goal {\sf parse(KW, T)}, where {\sf KW} is assumed
bound to {\sf [kim,walks]} (although the binding is
actually immaterial, as no step in
this refutation instantiates this variable).
The selection rule expresses a ``preference''
for goals with certain arguments instantiated.
If there is a literal of the form {\sf wf(}$T,C${\sf)} with $T$
instantiated to a non-variable then the left-most such literal is selected,
otherwise if there is a literal of the form {\sf y(}$T, S0, S${\sf)}
with $S0$ instantiated to a non-variable then the left-most such
literal is selected,
otherwise the left-most literal is selected.
The selected literal is underlined, and
the new literals introduced by each reduction are inserted
to the left of the old literals.

\begin{figure}
\beginprolog \sf
{\rm (1)} \underline{parse(KW,T)}.
{\rm (2)} wf(T,s), \underline{y(T,KW,[])}.
{\rm (3)} \underline{wf(\uline/[T1,T2],s)}, y(T1,KW,S1), y(T2,S1,[]).
{\rm (4)} wf(T1,np), wf(T2,vp), \underline{y(T1,KW,S1)}, $\ldots$
{\rm (5)} y(T3,KW,S3), y(T4,S3,S1), \underline{wf(\uline/[T3,T4],np)}, $\ldots$
{\rm (6)} wf(T3,np), wf(T4,n), \underline{y(T3,KW,S3)}, $\ldots$
{\rm (7)} y(T5,KW,S5), y(T6,S5,S3), \underline{wf(\uline/[T5,T6],np)}, $\ldots$
    $\ldots$
\endprolog \caption{An infinite SLD refutation, despite co-routining}
\label{fig:sld}
\end{figure}

Note that the {\sf y} and {\sf wf} literals resolved
at steps (6) and (7) in Figure~\ref{fig:sld}
are both children of and variants of
the literals resolved at steps (5) and (6).
This sequence of resolution steps can
be iterated an arbitrary number of times.
It is a manifestation of the left recursion
in the well-formedness constraint {\sf wf}.

One standard technique for dealing with such left-recursion
is memoization~\cite{pw:ed,ps:pnlp}.
But there are two related problems in applying the standard
logic programming memoization techniques to this problem.

First,
because the standard methods
memoize and evaluate at the level
of an individual literal,
the granularity at which they apply memoization is
is to small.
As noted above, in general individual
{\sf wf} or {\sf y} literal can have an infinite
number of answer substitutions.
The lemma table proof procedure circumvents
this problem by memoizing {\em conjunctions}
of literals (in this example, a conjunction
of {\sf wf} and {\em y} literals which has
only a finite number of subsumption-incomparable
valid instances).

The second problem is that the standard memoization
techniques restrict the order in which literals
can be resolved.
In general, these restrictions prevent the
``goal delaying'' required to co-routine
among several constraints.
The lemma table proof procedure lifts this
restriction by allowing arbitrary selection
rules.

\section{The Lemma Table proof procedure}
Like the Earley Deduction and the OLDT proof procedures,
the Lemma Table proof procedure maintains a lemma table
that records
goals and their corresponding solutions.
After a goal has been entered into the lemma table,
other occurences of instances of that goal can be reduced
by the solutions from the lemma table instead of the
original program
clauses.

We now turn to a formal presentation of the Lemma Table proof procedure.
In what follows,
lower-case letters are used for variables that range over atoms.
Upper-case letters are used for variables that
range over {\em goals},
which are sets of atoms.  Goals are interpreted
conjunctively; a goal is satisfied iff all of
its members are.

A goal $ G $ {\em subsumes} a goal $ G' $ iff
there is some substitution $\theta$ such that
$G' = G \theta$.  (Note that m.g.u.'s for sets
of goals are in general not unique even up to renaming).

The ``informational units'' manipulated by the lemma
table proof procedure are called generalized clauses.
A {\em generalized clause} is a pair of goals,
and is written $ \clause{G_1}{G_2} $.
$G_1$ is called the {\em head} of the clause and $G_2$
is called the {\em body}.  Both the head and body
are interpreted conjunctively;
i.e., $ \clause{G_1}{G_2} $ should be read as
``if each of the $G_2$ are true, then all of the $G_1$ are true''.
A generalized clause has a natural interpretation as
a goal subject to constraints:
$G_1$ is true in any interpretation which satisfies
the constraints expressed by $G_2$.

A {\em lemma table} is a set of {\em table entries}
of the form $\langle G, T, S \rangle$, where
\begin{enumerate}
 \item $G$ is a goal (this entry is called a table entry {\em for} $G$),
 \item $T$ is a lemma tree (see below), and\
 \item $S$ is a sequence of clauses, called the
  {\em solution list} for this entry.
\end{enumerate}

A {\em lemma tree} is a tree constructed by the
algorithm described below.  Its nodes have two
labels. These are
\begin{enumerate}
\item a clause $\clause{A}{B}$, called the {\em clause labelling} of
the node, and
\item an optional tag, which when present is one of
$\solution$, $\program(b)$ for some $b \in B$,
or $\lookup(B',p)$
where $ \emptyset \subset B' \subseteq B$
and $p$ is either the null pointer $\nil$
or a pointer into a solution list of a table entry
for some $G$ that subsumes $B'$.
\end{enumerate}

Untagged nodes are nodes that have not yet
been processed.  All nodes are untagged when they are
created,
and they are assigned a tag as they are processed.
The tags indicate which kind of resolution has been
applied to this clause.
A node tagged $\program(b)$ is resolved
against the clauses defining $b$ in the program.
A node tagged $\lookup(B',p)$ is resolved against
the instances of a table entry $E$ for some goal
that subsumes $B'$; the pointer $p$ keeps track
of how many of the solutions from $E$ have been
inserted under this node (just as in OLDT resolution).
Finally, a node tagged $\solution$ is not resolved,
rather its clause labelling is added to the solution
list for this table entry.

Just as SLD resolution is controlled by a selection rule
that determines which literal will be reduced next,
the Lemma Table proof procedure is controlled by a control
rule $R$ which determines the next goal (if any) to be reduced
and the manner of its reduction.

More precisely, $R$ must tag a node with clause labelling $\clause{A}{B}$
with a tag that is either $\solution$,
$\program(b)$ for some $b \in B$, or
$\lookup(B',\nil)$ such that $ \emptyset \subset B' \subseteq B$.
Further,
$R$ must tag the root node of every lemma tree with the
tag $\program(b)$ for some $b$
(this ensures that some program reductions are performed
in every lemma tree, and hence that a lemma table entry cannot
be used to reduce itself vacuously).

Finally, as in OLDT resolution, the Lemma Table proof procedure
allows a user-specified {\em abstraction operation}
$\alpha$
that maps goals to goals such that
$\alpha(G)$ subsumes $G$ for all goals $G$.
This is used to generalize the goals in the same way as
the term-depth abstraction operation in OLDT resolution,
which it generalizes.

The Lemma Table proof procedure can now be presented.

\begin{description}
\item[Input:] A non-empty goal $G$, a program $P$,
an abstraction operation $\alpha$,
and a control rule $R$.
\item[Output:] A set $\gamma$ of clauses of the
form $\clause{G'}{C}$, where $G'$ is an instance of $G$.
\item[Algorithm:] Create a lemma table with one table entry
$\langle G, T, \emptylist \rangle$, where $T$ contains a single untagged
node with the clause labelling $\clause{G}{G}$.
Then repeat the following operations until no operation applies.
Finally, return the solution list from the table entry for $G$.

The operations are as follows.

\begin{description}
\item[Prediction]
Let $v$ be an untagged node in a lemma tree
$T$ of table entry $\langle G, T, S \rangle$, and let
$v$'s clause labelling be $\clause{A}{B}$.
Apply the rule $R$ to $v$, and perform the action specified below
depending on the form of the tag $R$ assigned to $v$.
\begin{description}
\item[$\solution:$] Add $\clause{A}{B}$ to the end of
the solution list $S$.
\item[$\program(b):$] Let $B' = B - \{ b \}$.  Then for each
clause $\clause{b'}{C}$ in $P$ such that $b$ and $b'$ unify with a
m.g.u.\ $\theta$, create an untagged child node $v'$ of $v$ labelled
$(\clause{A}{B' \cup C})\theta$.
\item[$\lookup(B',\nil) :$]  The action in this case depends on
whether there already is a table entry
$\langle G', T', S' \rangle$ for some $G'$ that subsumes $B'$.
If there is, set the pointer in the tag to the start of the
sequence $S'$.
If there is not, create a new table entry
$\langle \alpha(B'), T'', \emptylist \rangle$,
where $T''$ contains a single untagged node
with clause labelling $\clause{\alpha(B')}{\alpha(B')}$.
Set the pointer in $v$'s tag to point to the empty solution
list of this new table entry.
\end{description}
\item[Completion]
Let $v$ be a node with clause labelling $\clause{A}{B}$
and tagged $\lookup(B',p)$ such that $p$ points
to a non-null portion $S'$ of a solution list of some
table entry.
Then advance $p$ over
the first element $\clause{B''}{C}$ of $S'$ to point to the remainder
of $S'$.  Further, if $B'$ and $B''$ unify with m.g.u.\ $\theta$,
then add a new untagged child node to $v$ labelled
$(\clause{A}{(B - B') \cup C})\theta$.
\end{description}
\end{description}

\begin{figure}
\picture 6.11in by 7.46in ('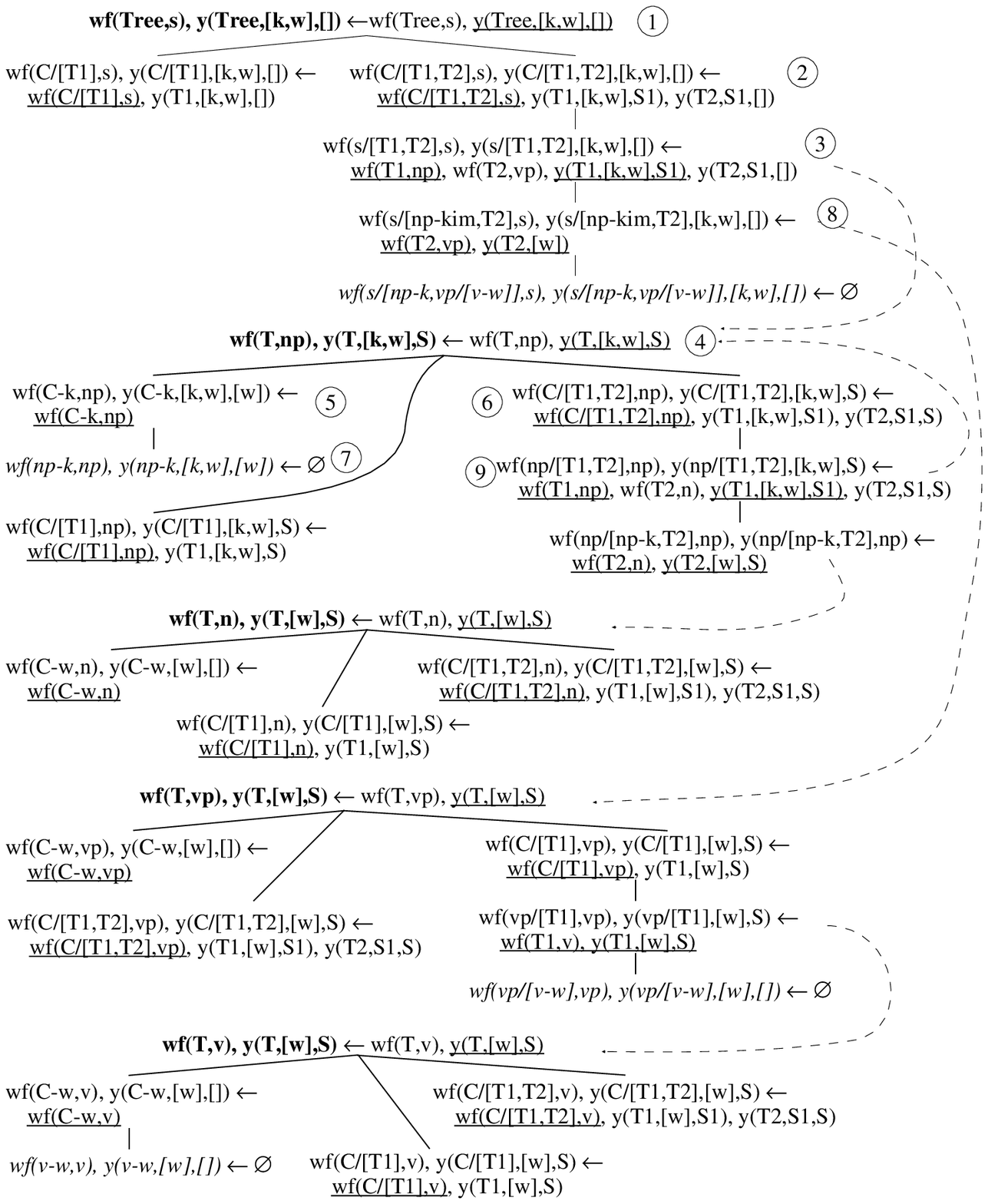')
\caption{A lemma table for {\sf wf(Tree,s), y(Tree,[kim,walks],[])}} %
\label{fig:lemmas}
\end{figure}

It may help to consider an example based on the program in
Figure~\ref{fig:grammar}.
The computation rule $R$ used is the following.
Let $v$ be a node in a lemma tree and let $\clause{A}{B}$ be
its clause label.
If $v$ is the root of a lemma tree and $B$ contains
a literal of the form ${\sf y(}T,S0,S{\sf)}$ then
$R(v) = \program({\sf y(}T,S0,S{\sf)})$.
If $B$ is empty then $R(v) = \solution$ (no other tagging is possible
for such nodes).
If $B$ contains a literal of the form ${\sf wf(}T,C{\sf)}$
where $T$ is a non-variable
then $R(v) = \program({\sf wf(}T,C{\sf)})$ (there is never more
than one such literal).
Otherwise, if $B$ contains two literals of the form
${\sf wf(}T,C{\sf)}, {\sf y(}T,S0,S{\sf )}$ where $S0$ is a
non-variable, then
$R(v) = \lookup(\{ {\sf wf(}T,C{\sf)}, {\sf y(}T,S0,S{\sf )} \},\nil)$.
These four cases exhaust all of the node labelling encountered
in the example.\footnote{
When used with a program encoding a context-free grammars in
the manner of Figure~\ref{fig:grammar},
the Lemma Table proof procedure with
the control rule $R$ simulates Earley's CFG
parsing algorithm~\cite{e:ep}.
The operations in the Lemma Table proof procedure
are named after the corresponding operations of Earley's algorithm.}

Figure~\ref{fig:lemmas} depicts the completed lemma table constructed
using the rule $R$ for the goal
{\sf wf(Tree,s), y(Tree,[kim,walks],[])}.
To save space, {\sf kim} and {\sf walks} are abbreviated
to {\sf k} and {\sf w} respectively.
Only the tree from each table entry is shown
because the other components of the entry
can be read off the tree.
The goal of each table entry is the head of the clause labelling
its root clause, and is shown in bold face.
Solution nodes are shown in italic face.
Lookup nodes appear with a dashed line pointing
to the table entry used to reduce them.

The first few steps of the proof procedure are the following;
each step corresponds to the circled node of the same number.
\begin{enumerate}
\item[(1)]
The root node of the first table entry's tree
restates the goal to be proven.
Informally, this node searches for an S located
at the beginning of the utterance.
The literal
{\sf y(Tree,[kim,walks],[])}
is selected for program reduction.
This reduction produces two child nodes, one of
which ``dies'' in the next step because there are
no matching program nodes.
\item[(2)]
The reduction (1) partially instantiates
the parse tree {\sf Tree}.
The literal {\sf wf(C/[T1,T2])} is selected for program
reduction.
\item[(3)]
This produces a clause body that contains literals that
refer to the subtree {\sf T1} and literals that
refer to the subtree {\sf T2}.  The computation rule
in effect partitions the literals and
selects those that refer to the subtree {\sf T1}
(because they are associated with an instantiated
left string argument).
\item[(4)]
A new table entry is created for the literals selected
in (3).
Informally, this entry searches for an NP located
at the beginning of the utterance.
Because this node is a root node, the
literal {\sf y(T,[kim,walks],S)} is selected for
program expansion.
\item[(5--6)]
The literals with predicate {\sf wf} are selected
for program expansion.
\item[(7)]
The body of this node's clause label is empty, so
it's label is added to the solutions list of the
table entry.
Informally, this node corresponds to the string
{\sf [kim]} having been recognized as an NP.
\item[(8)]
The solution found in (7) is incorporated into
the tree beneath (3).
A new table entry is generated to search for
a VP spanning the string {\sf [walks]}.
\item[(9)]
Just as in (3), the literals in the body of this
clause's label refer to two distinct subtrees, and
as before
the computation rule selects the literals that
refer to {\sf T1}.
However there is already a table entry (4) for the
selected goal, so a new table entry is not created.
The solutions already found for (4) generate a child
node to search for an N beginning at {\sf walks}.
No solutions are found for this search.
\end{enumerate}

\section{Soundness and Completeness}
This section demonstrates the soundness and completness of
the lemma table proof procedure.  Soundness is straight-forward,
but completeness is more complex to prove.
The completeness proof relies on the notion of
an unfolding of a lemma tree,
in which the $\lookup$ nodes of a lemma tree are systematically
replaced with the tree that they point to.
In the limit, the resulting tree can be viewed
a kind of SLD proof tree, and completeness follows
from the completeness of SLD resolution.

\begin{Theorem}[Soundness]
If the output of lemma table proof
procedure contains a clause $\clause{G}{C}$, then
$ P \models \, C \rightarrow G$.
\end{Theorem}
{\bf Proof:}
Each of the clause labels on lemma tree nodes
is either a tautology or derived by resolving
other clause labels and program clauses.
Soundness follows by induction on the number
of steps taken by the proof procedure.
\eop

As might be expected, the completeness proof
is much longer than the soundness proof.
For space reasons it is only sketched here.

For the completeness proof we assume
that the control rule $R$ is such
that the output $\gamma$ of the lemma table proof procedure
contains only clauses with empty bodies.
This is reasonable in the current context, because
non-empty clause bodies correspond to goals
that have not been completely reduced.

Then completeness follows if for all $P$, $G$ and $\sigma$, if
$ P \models G \sigma$
then there is an instance $G'$ on the solution list
that subsumes $G \sigma$.

Further, without loss of generality
the abstraction operation $\alpha$
is assumed to be the identity function on goals,
since
if $\alpha(G(\vec{t})) = G(\vec{t'}) $,
the goal $G(\vec{t})$ can be replaced
with the equivalent
$ G(\vec{t'}) \cup \{ \vec{t'} = \vec{t} \} $
and $\alpha$ taken to be the identity function.

Now, it is a corollary of the
Switching Lemma~\cite[pages~45--47]{l:lp}
that if $ P \models G \sigma$ then
there is an $n$ such that for any computation rule
there is an SLD refutation of $G$ of length $n$ whose
computed answer substitution $\theta$ subsumes $\sigma$.
We show that if $\theta$ is a computed answer substitution
for an SLD derivation of length $n$ then there is
a node tagged $\solution$ and labelled
$\clause{G \theta}{\emptyset}$ in
lemma tree $T$ for the top-level
goal.

First, a well-formedness condition on lemma
trees is introduced. Every lemma tree in a lemma
table at the termination of the lemma table proof procedure
is well-formed.  Well-formedness and the set of nodes tagged
$\solution$ are preserved
under an abstract operation on lemma trees called expansion.
The expansion of a lemma tree is
the tree obtained by replacing each node tagged
$\lookup(B',p)$ with the lemma
tree in the table entry pointed to by $p$.

Because expansions preserves well-formedness,
the lemma tree $T'$ resulting from $n$ iterated expansions
of the lemma tree $T$ for the top-level goal
is also well-formed.
Moreover, since the root node of every lemma tree is required
to be tagged $\program$,
all nodes in $T'$ within distance $n$ arcs
of the root will be tagged $\program$ or $\solution$.
This top part of $T'$ is isomorphic
to the top part of an SLD tree $T_s$ for $G$, so if $\theta$ is
a computed answer substitution for an SLD derivation in $T_s$
of length $n$ or less then there is a node tagged
$\solution$ and labelled $\clause{G \theta}{\emptyset}$
in $T'$, and hence $T$.
Since $n$ was arbitrary, every SLD refutation in $T_s$
has a corresponding node tagged $\solution$
in $T'$ and hence in $T$.

\section{Conclusion}
This paper generalizes standard memoization techniques
for logic programming to allow them to be used for
constraint logic programming.
The basic informational unit used in the Lemma Table
proof procedure is the generalized clause $\clause{G}{C}$.
Generalized clauses can be given a constraint interpretation
as ``any interpretation which satisfies the constraints $C$ also satisfies
$G$''.
The lemmas recorded in the lemma table
state how sets of literals are reduced to other sets of
literals.  Because the heads of the lemmas
consist of sets of literals rather than just
individual literals, the lemmas express properties of
systems of constraints rather than just individual constraints.
Because the solutions recorded in the lemma table
can contain unresolved constraints, it is possible
to pass constraints out of a lemma into the superordinate
computation.

In this paper $G$ and $C$ were taken to be sets of
literals and the constraints $C$ were defined
by Horn clauses.
In a more general setting, both $G$ and $C$ would
be permitted to contain constraints drawn from a
specialized constraint language not defined by
a Horn clause program.
H\"ohfeld and Smolka~\cite{hs:drcl} show how to extend SLD resolution
to allow general constraints over arbitrary domains.
Their elegant relational approach seems to be straight-forwardly applicable
to the Lemma Table proof procedure, and would actually
simplify its theoretical description
because equality (and unification) would be treated
in the constraint system.
Unification failure would then be a special case of
constraint unsatisfiability, and would be handled
by the ``optimization'' described by H\"ofeld and Smolka
that permits nodes labelled with clauses $\clause{G}{C}$
to be deleted if $C$ is unsatisfiable.

\end{document}